\documentclass{article}

\usepackage{microtype}
\usepackage{graphicx}
\usepackage{subfigure}
\usepackage{booktabs} 
\usepackage{kotex}

\usepackage{amsmath,amsfonts,bm}

\def\eqref#1{equation~\ref{#1}}

\def\1{\bm{1}}

\def\mA{{\bm{A}}}

\def\mD{{\bm{D}}}

\def\mI{{\bm{I}}}

\def\mP{{\bm{P}}}

\DeclareMathAlphabet{\mathsfit}{\encodingdefault}{\sfdefault}{m}{sl}
\SetMathAlphabet{\mathsfit}{bold}{\encodingdefault}{\sfdefault}{bx}{n}

\def\gG{{\mathcal{G}}}

\newcommand{\R}{\mathbb{R}}

\usepackage{hyperref}
\usepackage{url}
\usepackage{graphicx}
\usepackage{amsmath}
\usepackage{stmaryrd}
\usepackage{sidecap}
\usepackage{mathtools}
\usepackage[normalem]{ulem}
\useunder{\uline}{\ul}{}
\usepackage{ctable}
\usepackage{multirow}

\usepackage[accepted]{icml2020}

\icmltitlerunning{Graphs, Entities, and Step Mixture}

\begin{document}

\twocolumn[
\icmltitle{Graphs, Entities, and Step Mixture}



\icmlsetsymbol{equal}{*}

\begin{icmlauthorlist}
\icmlauthor{Kyuyong~Shin}{c}
\icmlauthor{Wonyoung~Shin}{s}
\icmlauthor{Jung-Woo~Ha}{c}
\icmlauthor{Sunyoung~Kwon}{c}
\end{icmlauthorlist}

\icmlaffiliation{c}{Clova AI Research, NAVER Corp.}
\icmlaffiliation{s}{Naver Shopping, NAVER Corp.}
\icmlcorrespondingauthor{Sunyoung Kwon}{sunny.kwon@navercorp.com}

\icmlkeywords{Machine Learning, ICML}

\vskip 0.3in
]


\printAffiliationsAndNotice{}  
\newcommand{\fix}{\marginpar{FIX}}
\newcommand{\new}{\marginpar{NEW}}

\newcommand{\prgname}{GESM}
\newcommand{\eg}{{\it e.g.}}
\newcommand{\ie}{{\it i.e.}}

\newcommand{\org}[1]{\textcolor{orange}{#1}}
\newcommand{\blue}[1]{\textcolor{blue}{#1}}

\newcommand{\vio}[1]{\textcolor{violet}{#1}}
\newcommand{\cyan}[1]{\textcolor{cyan}{#1}}

\begin{abstract}
Existing approaches for graph neural networks commonly suffer from the oversmoothing issue, regardless of how neighborhoods are aggregated. Most methods also focus on transductive scenarios for fixed graphs, leading to poor generalization for unseen graphs. To address these issues, we propose a new graph neural network that considers both edge-based neighborhood relationships and node-based entity features, i.e. \textbf{G}raph \textbf{E}ntities with \textbf{S}tep \textbf{M}ixture via \textit{random walk} (\prgname{}). \prgname{} employs a mixture of various steps through random walk to alleviate the oversmoothing problem, attention to dynamically reflect interrelations depending on node information, and structure-based regularization to enhance embedding representation. With intensive experiments, we show that the proposed \prgname{} achieves state-of-the-art or comparable performances on eight benchmark graph datasets comprising transductive and inductive learning tasks. Furthermore, we empirically demonstrate the significance of considering global information.
\end{abstract}
\section{Introduction}\label{sec:intro}
Graphs are universal data representations that exist in a wide variety of real-world problems, such as analyzing social networks~\citep{perozzi2014deepwalk, jia2017random}, forecasting traffic flow~\citep{manley2015estimating, yu2017spatio}, and recommending products based on personal preferences~\citep{page1999pagerank, kim2019tripartite}. Owing to breakthroughs in deep learning, recent graph neural networks (GNNs)~\citep{scarselli2008graph} have achieved considerable success on diverse graph problems by collectively aggregating information from graph structures~\citep{wang2018graphgan, xu2018representation, gao2019graph}. As a result, much research in recent years has focused on how to aggregate the feature representations of neighbor nodes so that the dependence of graphs is effectively utilized.

The majority of studies have predominantly depended on edges to aggregate the neighboring nodes' features. These edge-based methods are premised on the concept of relational inductive bias within graphs~\citep{battaglia2018relational}, which implies that two connected nodes have similar properties and are more likely to share the same label~\citep{kipf2016semi}. While this approach leverages graphs’ unique property of capturing structural relations, it appears less capable of generalizing to new or unseen graphs~\citep{wu2019comprehensive}.

To improve the neighborhood aggregation scheme, some studies have incorporated node information; they fully utilize node information and reduce the effects of structural (edge) information. A recent approach, graph attention networks (GAT), employs the attention mechanism so that weights used for neighborhood aggregation differ according to the feature of nodes~\citep{velivckovic2017graph}. This approach has yielded impressive performance and has shown promise in improving generalization for unseen graphs.

Regardless of neighborhood aggregation schemes, most methods, however, suffer from a common problem where neighborhood information is considered to a limited degree~\citep{klicpera2018predict}. For example, graph convolutional networks (GCNs)~\citep{kipf2016semi} only operate on data that are closely connected due to oversmoothing~\citep{li2018deeper}, which indicates the ``washing out" of remote nodes' features via averaging and becomes indistinguishable. Consequently, information becomes localized and access to global information is restricted~\citep{xu2018representation}, leading to poor performance on datasets in which only a small portion is labeled~\citep{li2018deeper}.

In order to address the aforementioned issues, we propose a novel method, \textbf{G}raph \textbf{E}ntities with \textbf{S}tep \textbf{M}ixture via \textit{random walk} (\prgname), which considers information from all nodes in the graph and can be generalized to new graphs by incorporating \textit{random walk} and \textit{attention}. \textit{Random walk} enables our model to be applicable to previously unseen graph structures, and a mixture of random walks alleviates the oversmoothing problem, allowing global information to be included during training. Hence, our method can be effective, particularly for nodes in the periphery or a sparsely labeled dataset. The \textit{attention} mechanism also advances our model by considering node information for aggregation. This enhances the generalizability of models to diverse graph structures. 

Despite the attention mechanism, it is likely that some homogeneous neighbor nodes are not still clustered closely on the embedding space. We employ a triplet loss-based regularization term~\citep{gordo2017end}, which enforces the homogeneous neighbor nodes to be closer.

To validate our approach, we conducted experiments on eight standard benchmark datasets: Cora, Citeseer, and Pubmed, which are well-known citation networks datsets, Coauthor CS, Coauthor Physics, Amazon Computers and Amazon Photo which are co-authorship and co-purchase datsets for transductive learning. We also conduct experiments on protein-protein interaction (PPI) dataset for inductive learning, in which test graphs remain unseen during training. In addition to these experiments, we verified whether our model uses information of remote nodes by reducing the percentage of labeled data. The experimental results demonstrate the consistently competitive performances of \prgname{} on all the datasets including transductive and inductive scenarios.

The key contributions of our approach are as follows:
\begin{itemize}
  \item We propose \textbf{G}raph \textbf{E}ntities with \textbf{S}tep \textbf{M}ixture via \textit{random walk} (\prgname{})\footnote[1]{The code is available on https://bit.ly/37bIb7W}, which incorporates \textbf{S}tep \textbf{M}ixture with bilinear pooling based attention and novel regularization technique.
  
  \item We experimentally demonstrate that our proposed model is consistently competitive compared to other models on eight benchmark datasets and is applicable to both transductive and inductive learning tasks. In addition, we show its effectiveness in the ability of global aggregation as the rate of labels decreases.

  \item We provide an in-depth analysis regarding the effects on performance and inference time as the propagation step increases, and confirm our superiority on oversmoothing issue.
\end{itemize}

\section{Preliminary}
\subsection{Node Classification Task}
Node classification is a task to classify the labels of the masked nodes by learning from the other nodes of given graphs such as citation, co-purchase relation, and protein-protein interaction networks. The node classification can be categorized into \emph{transductive learning} and \emph{inductive learning} tasks depending on unseen graphs.

For transductive learning to handle a single fixed graph, it is important to learn the structural representation of the graph nodes. Thus, GCN~\citep{kipf2016semi} and HGCN~\citep{chami2019hyperbolic}, which are state-of-the-art models of transductive learning, focus on how effectively 
to handle the graph structural representation. However, these models learn from spectral-domain or hyperbolic-space by limiting the graph structure to only one fixed graph~\citep{hamilton2017inductive}, thus leading to poor generalization on unseen graphs.

For inductive learning tasks to deal with multiple graphs, existing models such as GraphSAGE~\citep{hamilton2017inductive} and GAT~\citep{velivckovic2017graph} employed methods for a local range of aggregations by neighborhood node sampling and actively utilizing node embedding. However, in the case of sampling, there are issues of which node and how many nodes to sample. In the case of node embedding, excessively high computation time and loses the importance of structure information still remain challenging.

In addition, most existing approaches for graph representation learning commonly suffer from the oversmoothing issue. Despite the efforts of recent work including JK-Net~\citep{xu2018representation} and APPNP~\citep{klicpera2018predict}, they cannot completely solve the oversmoothing issue. Since APPNP employes a simple sum of Neumann series~\citep{gleich2016seeded}, it cannot adjust global and local aggregation scheme. In the case of JK-Net, global aggregation does not work effectively as shown in~\figurename~\ref{fig:test2}. 

\subsection{Random Walks}
\begin{figure}
  \centering
  \includegraphics[width=\columnwidth]{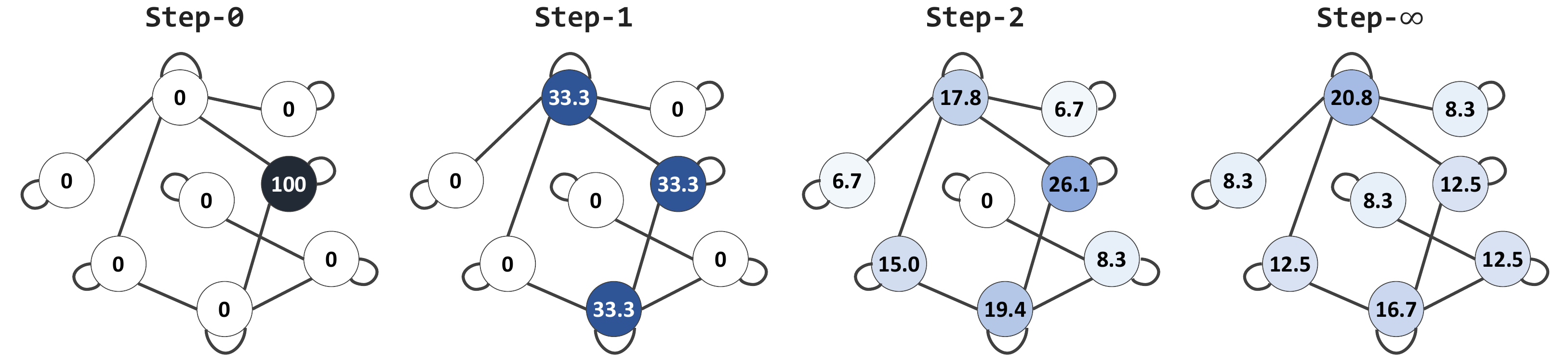}
  \caption{ Random walk propagation procedure. From left to right are step-0, step-1, step-2, and step-infinite. The values in each node indicate the distribution of a random walk. In the leftmost picture, only the starting node has a value of 100, and all other nodes are initialized to zero. As the number of steps increases, values spread throughout the graph and converge to some extent.} 
  \label{fig:random walks}
\end{figure}
\begin{figure*}
\centering
\subfigure[Traditional global aggregation scheme
]{\label{fig:a}\includegraphics[width=0.32\textwidth]{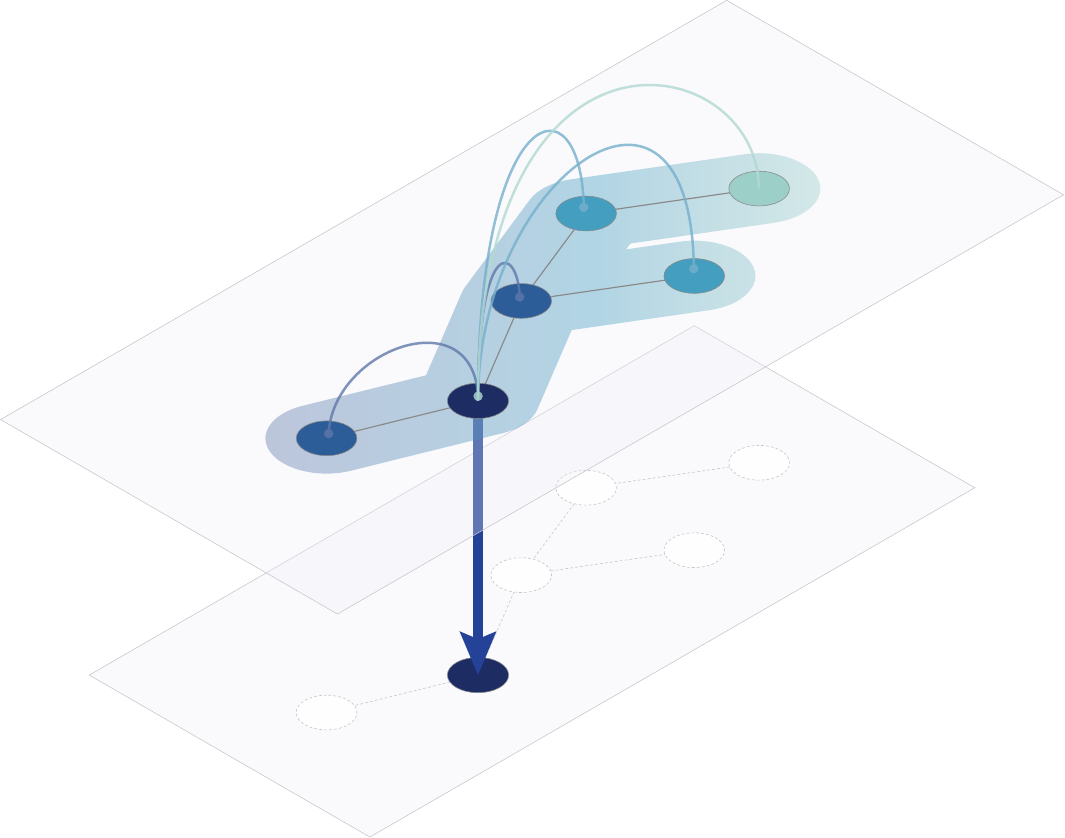}}
\subfigure[Our step-mixture scheme]{\label{fig:b}\includegraphics[width=0.32\textwidth]{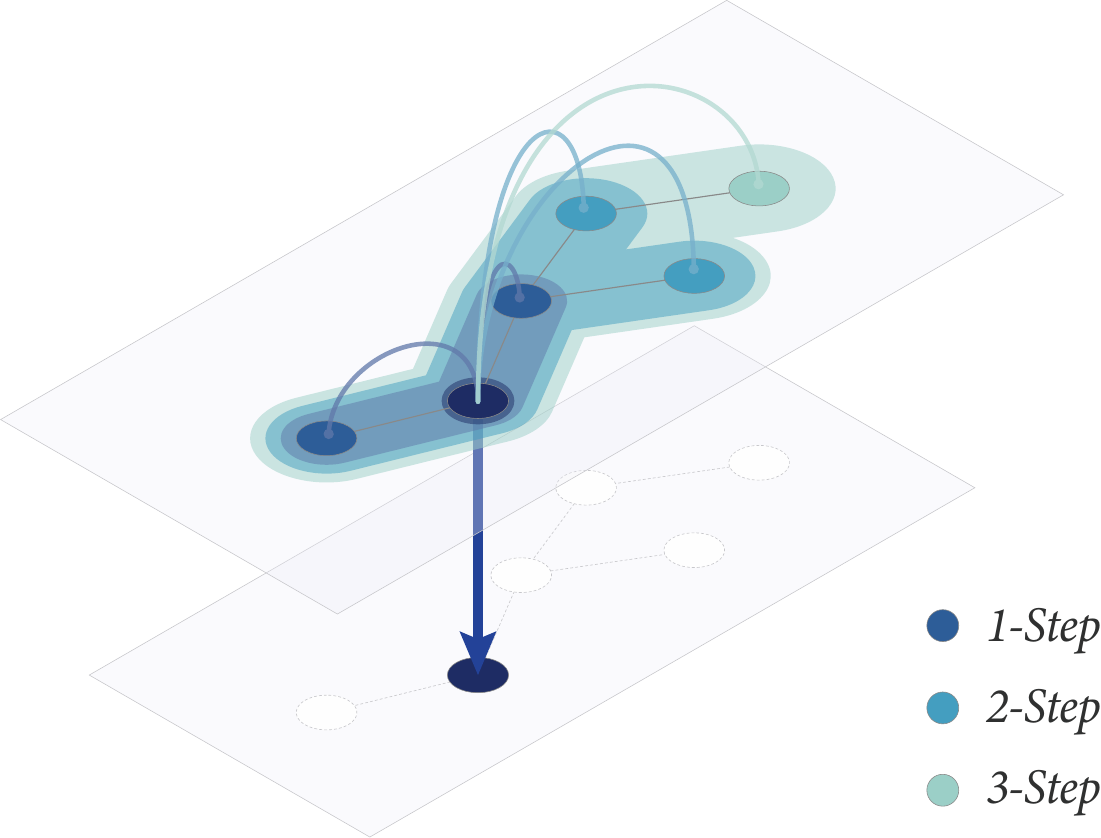}}
\caption{Conceptual scheme of neighborhood aggregation for three steps in conventional graph neural networks (a) and our method which uses mixture of random walks (b).}
\label{fig:step-mixture}
\end{figure*}
Random walk, which is a widely used method in graph theory, mathematically models how node information propagates throughout the graph. As shown in~\figurename~\ref{fig:random walks}, random walk refers to randomly moving to neighbor nodes from the starting node in a graph. For a given graph, the transition matrix $\mP$, which describes the probabilities of transition, can be formulated as follows:
\begin{equation} \label{eq1-2}
\mP=\mA\mD^{-1}
\end{equation}
where $\mA$ denotes the adjacency matrix of the graph, and $\mD$ the diagonal matrix with a degree of nodes. The probability of moving from one node to any of its neighbors is equal, and the sum of the probabilities of moving to a neighboring node adds up to one. 

Let $u^t$ be the distribution of the random walk at step $t$ ($u^0$ represents the starting distribution).
The $t$ step random walk distribution is equal to multiplying $\mP$, the transition matrix, $t$ times. In other words,
\begin{equation} \label{eq1-4}
\begin{split}
u^1 & =\mP u^0 \\
u^{t} & =\mP u^{t-1}=\mP^t u^0.
\end{split}
\end{equation} 
The entries of the transition matrix are all positive numbers, and each column sums up to one, indicating that $\mP$ is a matrix form of the Markov chain with steady-state. One of the eigenvalues is equal to 1, and its eigenvector is a steady-state~\citep{strang1993introduction}. Therefore, even if the transition matrix is infinitely multiplied, convergence is guaranteed. 

\subsection{Attention}
The attention mechanism was introduced in sequence-to-sequence modeling to solve long-term dependency problems that occur in machine translation~\citep{bahdanau2014neural}. The key idea of attention is allowing the model to learn and focus on what is important by examining features of the hidden layer. In the case of GNNs~\citep{scarselli2008graph}, attention mechanism has been used to give different importance to neighboring nodes depending on their informational relevance to a center node. During the propagation process, node features are given more emphasis than structural information (edges). Consequently, using attention is advantageous for training and testing graphs with different node features even in the same structures (edges). GATs~\citep{velivckovic2017graph} achieved state-of-the-art performance by using the attention mechanism, which is based on concatenation operation between node information.

Given the many benefits of attention, we incorporate the attention mechanism to our model, which is based on bilinear pooling to fully utilize node information and fine-grained relevance. Our attention mechanism enables different weights to neighboring nodes by considering interactions. Combining attention with mixture-step random walk allows our model to adaptively highlight features with salient information in a global scope.

\setlength{\tabcolsep}{8pt}
\ctable[
    caption = {Summary of datasets used in the experiments.},
    label = tab:data,
    star,
      doinside=\small
]{clrrrrrrrr}{
\tnote[$\ast$] {Transductive learning datasets consist of one graph and use a subset of the graph for training. }
\tnote[$\dagger$]{Inductive learning datasets consist of many graphs and use a few graphs for training and unseen graphs for testing.
}
}{
\toprule
  \multirow{2}{*}{Type}  & \multirow{2}{*}{Datasets} & \multicolumn{4}{c}{Nodes (Graphs)}  & \multirow{2}{*}{Features} & \multirow{2}{*}{Edges} & \multirow{2}{*}{Classes}  &  \multirow{2}{*}{Label rate} \\ 
    \cmidrule(l){3-6}
  &   & Total & Training & Validation & Test & & & \\
\midrule
\multirow{3}{*}{T$^\ast$} & Cora     & 2,708 & 140 & 500 & 1,000          & 1,433              & 5,429           & 7     & 5.1\%               \\
 & Citeseer & 3,327    & 120 & 500 & 1,000       & 3,703              & 4,732           & 6     & 3.6\%               \\
 & Pubmed   & 19,717      & 60 & 500 & 1,000    & 500               & 44,338          & 3    & 0.3\%               \\
 \midrule
I$^\dagger$                   & PPI      & 56,944   & 44,906 & 6,514 & 5,524       & 50                & 8,187           & 121    & -                \\
& & (24) & (20) & (2) & (2) &  & &  &  \\
\bottomrule
}

\section{Graph Entity and Step Mixture (GESM)}
First, we define the notations used in this paper for convenience. Let $\gG = (V, E)$ be a graph, where $V$ and $E$ denote the sets of nodes and edges, respectively. Nodes are represented as a feature matrix $X \in \R^{n \times f}$, where $n$ and $f$ respectively denote the number of nodes and the input dimension per node. A label matrix is $Y \in \R^{n \times c}$ with the number of classes $c$, and a learnable weight matrix is denoted by $W$. The adjacency matrix of graph $\gG$ is represented as $\mA \in \R^{n \times n}$. The addition of self-loops to the adjacency matrix is $\widetilde{\mA} = \mA$ + $\mI_n$, and the column normalized matrix of $\widetilde{\mA}$ is  $\hat{\tilde{\mA}} = \widetilde{\mA}\mD^{-1}$ with $\hat{\tilde{\mA}}^0 = \mI_n$. 

\subsection{Step Mixture for Avoiding Oversmoothing}
Most graph neural networks suffer from the oversmoothing issue. Although JK-Net~\citep{xu2018representation} tried to handle oversmoothing by utilizing GCN blocks with mulitple propagation, it could not completely resolve the issue as shown in~\figurename~\ref{fig:test2}.  Unlike JK-Net, we explicitly separate the node embedding and the propagation process by employing a mixture of multiple random walk steps. This step mixture approach allows our model to alleviate the oversmoothing issue along with localized aggregation.

Our method has a simple structure that is composed of three stages. First, input $X$ passes
through a fully connected layer with a nonlinear activation and generate embedding node feature $Z=\sigma(XW)$. Second, $Z$ is multiplied by a normalized adjacency matrix $\hat{\tilde{\mA}}$ for each random walk step that is to be considered. The first and the second stages show node embedding and propagation processes are separated. Finally, the concatenated result of each step $f_{cat}$ is passing through the prediction layer. The entire propagation process can be formulated as:
\newcommand*\concat{\mathbin{\|}}
\begin{equation} \label{eq6}
f_{cat}=\bigparallel_{k=0}^s \hat{\tilde{\mA}}^{k}Z,
\end{equation}
where $\bigparallel$ is the concatenation operation, $s$ is the maximum number of steps considered for aggregation, and $\hat{\tilde{\mA}}^{k}$ is the normalized adjacency matrix $\hat{\tilde{\mA}}$ multiplied $k$ times. As can be seen from \equationautorefname~\ref{eq6}, weights are shared across nodes. 

In our method, the adjacency matrix $\hat{\tilde{\mA}}$ is an asymmetric matrix, which is generated by \textit{random walks} and flexible to arbitrary graphs. On the other hand, prior methods such as JK-Net~\citep{xu2018representation} and MixHop~\citep{abu2019mixhop}, use a symmetric Laplacian adjacency matrix, which limits graph structures to given fixed graphs. 

For the concatenation operation, localized sub-graphs are concatenated with global graphs, which allows the neural network to adaptively select global and local information through learning (see~\figurename~\ref{fig:step-mixture}). While traditional graph convolution methods consider  aggregated information within three steps by $\mA(\mA(\mA XW^{(0)})W^{(1)})W^{(2)}$, our method can take all previous aggregations into account by $(\mA^{0}Z \mid \mA^{1}Z \mid \mA^{2}Z \mid \mA^{3}Z$).

\subsection{Neighborhood Interaction-based Attention}
For more sophisticated design of node embedding $Z$, we adopt bilinear pooling-based neighborhood interaction as attention mechanism so that node information is relatively emphasized for aggregation. We simply replace $Z$ with the attention feature denoted by $H_{\textrm{multi}}$ as \equationautorefname s~\ref{eq6} and \ref{eq8},
\begin{equation} \label{eq8}
\centering
output=\text{softmax}(
(\bigparallel_{k=0}^s \hat{\tilde{\mA}}^{k}H_{\textrm{multi}})W)).
\end{equation}
As described in~\equationautorefname~\ref{eq7}, we employ multi-head attention, where $H_{\textrm{multi}}$ and $\alpha _i$ denote the concatenation of $m$ attention layers and the $i$-th attention coefficient. We only compute $\alpha$ for nodes $j \in \mathcal{N}_{i}$, the neighborhood of node $i$, to maintain the structure representation of the graph. The attention coefficients $\alpha$ is calculated by sum of outer product between encoding vectors of node $i$ and its neighbor node $j$:
\begin{equation} \label{eq7}
\centering
H_{\textrm{multi}}=\bigparallel_{i=1}^m \sigma(\alpha_{i}Z_{i})),
\end{equation}
\begin{equation} \label{eqatt}
\centering
\alpha_{i}=\text{softmax}_{j}(\sum e_{i}\otimes e_{j}),
\end{equation}
where $e$ is a hadamard-product of node embedding $Z$ and weight matrix $W$, $e=Z\odot W$.

By incorporating attention to our base model, we can avoid or ignore noisy parts of the graph, providing a guide for random walk~\citep{lee2018attention}. Utilizing attention can also improve combinatorial generalization for inductive learning, where training and testing graphs are completely different. In particular, datasets with the same structure but different node information can benefit from our method because these datasets can only be distinguished by node information. Focusing on node features for aggregation can thus provide more reliable results in inductive learning.
\setlength{\tabcolsep}{6pt}
\ctable[
    caption = {Experimental results on the public benchmark datasets. Evaluation metrics on transductive and inductive learning datasets are classification accuracy (\%) and F1-score, respectively. Top-3 results for each column are highlighted in bold, and top-1 values are underlined.
},
    label = tab:result,
    star,
     doinside = \footnotesize
]{lcccc}{
\tnote[$\ast$] {Best experimental results through our own implementation}
}{
\toprule
 & \multicolumn{3}{c}{Transductive} & Inductive \\
\cmidrule(l){2-4}  \cmidrule(l){5-5}
           & Cora                                                    & Citeseer                                               & Pubmed              & \multirow{2}{*}{PPI}                                \\
Method & public (5.1\%) & public (3.6\%) & public (0.3\%)  \\
\midrule
Cheby~\citep{defferrard2016convolutional} & 81.2   & 69.8  & 74.4 & -  \\
GCN~\citep{kipf2016semi}                  & 81.5  & 70.3  & 79.0  & -  \\
GraphSAGE~\citep{hamilton2017inductive}   & - & - & - & 0.612         \\
GAT~\citep{velivckovic2017graph}          & 83.0 & 72.5 & 79.0     & 0.973           \\
LGCN~\citep{gao2018large}  & \textbf{83.3}  & \textbf{\underline{73.0}} & 79.5 & 0.772 \\
JK-LSTM~\citep{xu2018representation}    & - & - & -  & \textbf{\underline{0.976}}   \\
AGNN~\citep{thekumparampil2018attention}  & 83.1 & 71.7  & 79.9    & -   \\
Union~\citep{li2018deeper}        & 80.5    & 65.7   & 78.3       & -       \\
$^\ast$APPNP~\citep{klicpera2018predict}  & 83.2 & 71.7  & 79.7    & -   \\
SGC~\citep{wu2019simplifying}  & 81.0 & 71.9  & 78.9    & -   \\
MixHop~\citep{abu2019mixhop}       & 81.9 & 71.4 & \textbf{\underline{80.8}} & -  \\
GWNN~\citep{xu2019graph}         & 82.8  & 71.7   & 79.1  & -   \\
AdaLNet~\citep{liao2019lanczosnet}      & 80.4  & 68.7 & 78.1  & -  \\ 
HGCN~\citep{chami2019hyperbolic}      & 79.9   & - & \textbf{80.3}  & -  \\ 
\midrule
\prgname{} (w/o att, reg)  & 82.8  & 71.7    & \textbf{80.3}   & 0.753 \\
\prgname{} (w/o reg)  & \textbf{84.4}  & \textbf{72.6}  & 80.1  & \textbf{\underline{0.976}}  \\
\prgname{}  & \textbf{\underline{84.5}}  & \textbf{72.7}  & \textbf{80.4}  & \textbf{0.974}  \\
\bottomrule
}

\subsection{Embedding Regularization}
Although our attention effectively filters out noisy nodes, 
there is still 
a possibility that the correct neighbor nodes may not be clustered in the process, as shown in~\figurename~\ref{fig:embedding}. Therefore, it is necessary for node embedding $Z$ to have a regularization that help to be clustered between neighbor nodes and to move away from irrelevant nodes by reflecting the graph structure. Our push and pull-based triplet regularization $R$ can be formulated as:
\begin{equation} \label{eq10}
\centering
R=\frac{1}{|S|}\sum_{\substack{p \in S \\ n \in S^{c}}}\left( \beta \cdot \text{Dis}(Z_c, Z_{p}) - (1-\beta) \cdot \text{Dis}(Z_c, Z_{n}) \right),
\end{equation}
where $S \subset E$ and its cardinality $|S|$ is number of samples, and $\beta$ denotes a weight for the distance of positive and negative. For a positive node $p$ represents the neighbor node of a center node $c$, and a negative node $n$, on the other hand, represents all nodes except the positive and center nodes. $\text{Dis}()$, as a distance function, we used a sigmoid of dot product, $\text{Dis}(Z_i, Z_j)=1-\text{sigmoid}(Z_{i}^{\mathsf{T}} Z_{j})$. 

Finally, our objective function $\mathcal{L}$ is $\mathcal{L}=J+R$, where $J$ denotes softmax cross-entropy loss between prediction label $output$ and target label, and $R$ denotes the regularizer.

\setlength{\tabcolsep}{10pt}
\ctable[
    caption = {Node classification results on datasets with low label rates. Top-3 results for each column are highlighted in bold and top-1 values are underlined.} ,
    label = tab:transductiveaccrandom,
    star,
      doinside=\small
]{lccccc}{
\tnote[$\ast$] {Best experimental results through our own implementation}
}{
\toprule
           & \multicolumn{2}{c}{Cora}                                     & \multicolumn{2}{c}{Citeseer}                                & \multicolumn{1}{c}{Pubmed}                                   \\
\cmidrule(l){2-3}  \cmidrule(l){4-5} \cmidrule(l){6-6}

Method  & 1\%                           & 3\%                          & 0.5\%                        & 1\%                          &  0.1\%                                                 \\ 
\midrule
Cheby~\citep{defferrard2016convolutional}        & 52.0                            & 70.8                         & 31.7                         & 42.8   & 51.2                \\
GCN~\citep{kipf2016semi}     & 62.3                          & 76.5                         & 43.6                         & 55.3  & 65.9                              \\
Union~\citep{li2018deeper}        & \textbf{69.9}                           & 78.5                         & 46.3                         & 59.1     & 70.7   \\
$^\ast$JK-GCN~\citep{xu2018representation}  & 65.1   & 76.8  & 37.1  & 55.3  & 71.1      \\
$^\ast$APPNP~\citep{klicpera2018predict}  & 67.6    & \textbf{80.8}  & 40.5  & 59.9  & 70.7       \\
$^\ast$SGC~\citep{wu2019simplifying}  & 64.2 & 77.2  & 41.0  & 58.1  & 71.7        \\
AdaLNet~\citep{liao2019lanczosnet}      & 67.5                          & 77.7                         & \textbf{\underline{53.8}}  & \textbf{\underline{63.3}}    & \textbf{72.8}                                                     \\ 
\midrule
\prgname{} (w/o att, reg)  & 68.2  &  \textbf{\underline{81.6}}                    & 45.6                             & 62.6  & \textbf{73.0}                       \\
\prgname{} (w/o reg)  & \textbf{70.5} & \textbf{81.2} & \textbf{53.2} & \textbf{62.7} & \textbf{\underline{73.8}}                                                     \\
\prgname{}  & \textbf{\underline{70.9}}  & 80.8  & \textbf{51.8}  & \textbf{63.0}  & \textbf{72.8} \\
\bottomrule
}

\section{Experiments}
\subsection{Datasets} 
\emph{\textbf{Transductive learning.}}
We utilize three benchmark datasets for node classification: Cora, Citeseer, and Pubmed~\citep{sen2008collective}. These three datasets are citation networks, in which the nodes represent documents and the edges correspond to citation links. The edge configuration is undirected, and the feature of each node consists of word representations of a document. Detailed statistics of the datasets are described in Table 1. 

For experiments on datasets with the public label rate, we follow the transductive experimental setup of~\cite{yang2016revisiting}. Although all of the nodes' feature vectors are accessible, only 20 node labels per class are used for training. Accordingly, 5.1\% for Cora, 3.6\% for Citeseer, and 0.3\% for Pubmed can be learned. In addition to experiments with public label rate settings, we conducted experiments using datasets where labels were randomly split into a smaller set for training. To check whether our model can propagate node information to the entire graph, we reduced the label rate of Cora to 3\% and 1\%, Citeseer to 1\% and 0.5\%, Pubmed to 0.1\%, and followed the experimental settings of ~\cite{li2018deeper} for these datasets with low label rates. For all experiments, we report the results using 1,000 test nodes and use 500 validation nodes. 

\emph{\textbf{Inductive learning.}}
We use the protein-protein interaction PPI dataset~\citep{zitnik2017predicting} ,which is preprocessed by~\cite{velivckovic2017graph}. As detailed in Table 1, the PPI dataset consists of 24 different graphs, where 20 graphs are used for training, 2 for validation, and 2 for testing. The test set remains completely unobserved during training. Each node is multi-labeled with 121 labels and 50 features regarding gene sets and immunological signatures.

\subsection{Experimental Setup} 
Regarding the hyperparameters of our transductive learning models, we set the dropout probability such that 0.7 and number of multi-head $m=8$. The size of the hidden layer $h \in \{64, 512\}$ and the maximum number of steps used for aggregation $s \in \{5, 15\}$ were adjusted for each dataset. We trained for a maximum of 300 epochs with L2 regularization $\lambda \in \{0.003, 0.0005\}$, triplet regularization $\beta \in \{0.5, 1.0\}$ and learning rate $lr \in \{0.003, 0.0008\}$. We report the average classification accuracy of 20 runs. 

For inductive learning, we set the size of hidden layer $h=256$, number of steps $s=3$, multi-head attention $m=8$, and $\beta=1.0$ for~\prgname{}. L2 regularization and dropout were not used for inductive learning~\citep{velivckovic2017graph}. We trained our models for a maximum of 3,000 epochs with learning rate $lr=0.008$. The evaluation metric was the micro-F1 score, and we report the averaged results of 10 runs.

For the models, the nonlinearity function of the first fully connected layer was an exponential linear unit (ELU)~\citep{clevert2015fast}. Our models were initialized using Glorot initialization~\citep{glorot2010understanding} and were trained to minimize the cross-entropy loss using the Adam optimizer~\citep{kingma2014adam}. We employed an early stopping strategy based on the loss and accuracy of the validation sets, with a patience of 20 epochs for Cora and 100 epochs for others. All the experiments were performed using NAVER Smart Machine Learning (NSML) platform~\cite{kim2018nsml,sung2017nsml}.

\section{Results}
\subsection{Node classification}
\emph{\textbf{Results on benchmark datasets.}}
\tableautorefname~\ref{tab:result} summarizes the comparative evaluation experiments for transductive and inductive learning tasks. In general, not only are there a small number of methods that can perform on both transductive and inductive learning tasks, but the performance of such methods is not consistently high. Our methods, however, are ranked in the top-3 for every task, indicating that our method can be applied to any task with large predictive power.

For transductive learning tasks, the experimental results of our methods are higher than or equivalent to those of other methods. As can be identified from the table, our model \prgname{} (w/o reg), which suggest the importance of considering node information in the aggregation process, outperforms many existing baseline models. These results indicate the significance of considering both global and local information with attention mechanism. It can also be observed that \prgname{} yielded more stable results than \prgname{} (w/o reg), suggesting the importance of reconstructing structural node representation in the aggregation process.

For the inductive learning task, our proposed model \prgname{} (w/o reg) and \prgname{}, surpasses the results of GAT, despite the fact that GAT consists of more attention layers. These results for unseen graphs are in good agreement with results shown by~\citet{velivckovic2017graph}, in which enhancing the influence of node information improved generalization.

\emph{\textbf{Results on datasets with low label rates.}}
To demonstrate that our methods can consider global information, we experimented on sparse datasets with low label rates of transductive learning datasets. As indicated in \tableautorefname~\ref{tab:transductiveaccrandom}, our models show remarkable performance even on the dataset with low label rates. In particular, we can further observe the superiority of our methods by inspecting \tableautorefname~\ref{tab:result} and~\ref{tab:transductiveaccrandom}, in which our methods trained on only 3\% of Cora dataset outperformed some other methods trained on 5.1\% of the data. Because our proposed model \prgname{} showed enhanced accuracy, it could be speculated that using a mixture of random walks played a key role in the experiments; the improved results can be explained by our methods adaptively selecting node information from local and global neighborhoods, and allowing peripheral nodes to receive information. 

\setlength{\tabcolsep}{5pt}
\ctable[
    caption = {Summary of datasets used in the additional experiments.
},
    label = tab:pitdescription,
     doinside = \footnotesize
]{lrrrrr}{
}{
\toprule & Classes    &  Features     & Nodes      & Edges \\
\midrule
Coauthor CS   &15 &6,805 &18,333 &100,227  \\
Coauthor Physics &5 &8,415 &34,493 &282,455   \\
Amazon Computers &10 &767 &13,381 &259,159  \\
Amazon Photo   &8 &745 &7,487 &126,530  \\
\bottomrule
}
\setlength{\tabcolsep}{6pt}
\ctable[
    caption = {Average test set accuracy and standard deviation over 100 random
train/validation/test splits with 20 runs. Top-3 results for each column are highlighted in bold, and top-1 values are underlined.},
    label = tab:pitfall,
    star,
      doinside=\small
]{lcccc}{
}{
\toprule & Coauthor CS    & Coauthor Physics     & Amazon Computers      & Amazon Photo  \\
\midrule
MLP     &88.3 $\pm$ 0.7 & 88.9 $\pm$ 1.1 & 44.9 $\pm$ 5.8 &69.6 $\pm$ 3.8 \\
LogReg    &86.4	$\pm$ 0.9 & 86.7 $\pm$ 1.5 & 64.1 $\pm$ 5.7 &73.0 $\pm$ 6.5 \\
GCN~\citep{kipf2016semi}        & 91.1 $\pm$ 0.5 & 92.8 $\pm$ 1.0 & \textbf{\underline{82.6}} $\pm$ 2.4 & \textbf{91.2} $\pm$ 1.2 \\
GraphSAGE~\citep{hamilton2017inductive} & 91.3 $\pm$ 2.8 & 93.0 $\pm$ 0.8 & \textbf{82.4} $\pm$ 1.8 & \textbf{\underline{91.4}} $\pm$ 1.3        \\
GAT~\citep{velivckovic2017graph}  & 90.5 $\pm$ 0.6 & 92.5 $\pm$ 0.9 & 78.0 $\pm$ 19.0 & 85.7 $\pm$ 20.3 \\
\midrule
\prgname{} (w/o att, reg) & \textbf{91.8}~$\pm$~0.4 & \textbf{93.3} $\pm$ 0.6 & 79.2 $\pm$ 2.0 & 89.3 $\pm$ 1.9  \\
\prgname{} (w/o reg)   & \textbf{\underline{91.5}} $\pm$ 0.5 & \textbf{93.4} $\pm$ 0.6 & 80.6 $\pm$ 2.1 & 89.8 $\pm$ 1.9   \\
\prgname{}   & \textbf{91.4} $\pm$ 0.5 & \textbf{\underline{93.5}} $\pm$ 0.8 & \textbf{80.8} $\pm$ 2.0 & \textbf{90.3} $\pm$ 2.1   \\
\bottomrule
}

\emph{\textbf{Experiments on other datasets for checking robustness.}}
For an in-depth verification of overfitting, we extended our experiments to four types of new node classification datasets.
Coauthor CS and Coauthor Physics are co-authorship graphs from the KDD Cup 2016 challenge\footnote[2] {https://kddcup2016.azurewebsites.net/}, in which nodes are authors, features represent the article keyword for each author's paper, and class labels indicate each author's most active research areas. Amazon Computers and Amazon Photo are co-purchase graphs of Amazon, where nodes represent the items, and edges indicate that items have been purchased together. The node features are bag-of-words of product reviews, and class labels represent product categories. Detailed statistics of the datasets are described in~\tableautorefname~\ref{tab:pitdescription} and we followed the experimental setup of~\cite{shchur2018pitfalls}.

We used the same values for each hyperparameter (unified size: 64, step size: 15, multi-head for GAT and GESM: 8) without tuning. The results in~\tableautorefname~\ref{tab:pitfall} prove that our proposed methods do not overfit to a particular dataset. Moreover, in comparison to GAT, the performance of GESM is more accurate and stable. 
\subsection{Model Analysis}\label{sec:model-analysis}
\emph{\textbf{Oversmoothing and Accuracy.}}
\begin{figure}
    \centering
    \includegraphics[width=0.75\columnwidth]{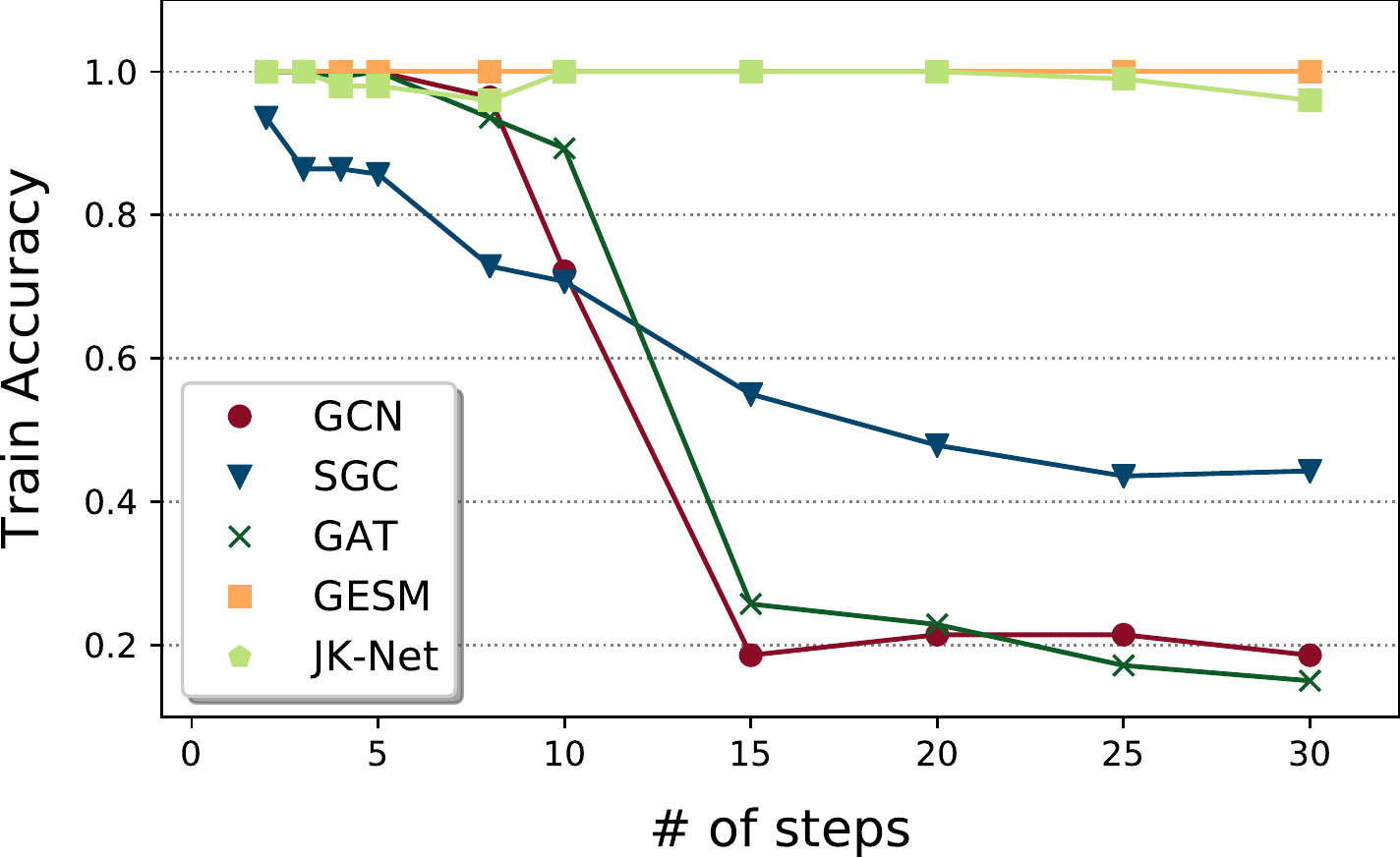}
    \caption{Training accuracy comparison according to number of propagation steps.}
    \label{fig:test1}
\end{figure}
As shown in~\figurename~\ref{fig:test1}, GCN~\citep{kipf2016semi}, SGC~\citep{wu2019simplifying}, and GAT~\citep{velivckovic2017graph} suffer from oversmoothing. GCN and GAT show severe degradation in accuracy after the 8th step; The accuracy of SGC does not drop as much as GCN and GAT but nevertheless gradually decreases as the step size increases. The proposed \prgname, unlike the others, maintains its performance without any degradation, because no rank loss~\citep{luan2019break} occurs and oversmoothing is overcome by step mixture.

\begin{figure}
    \centering
    \includegraphics[width=0.75\columnwidth]{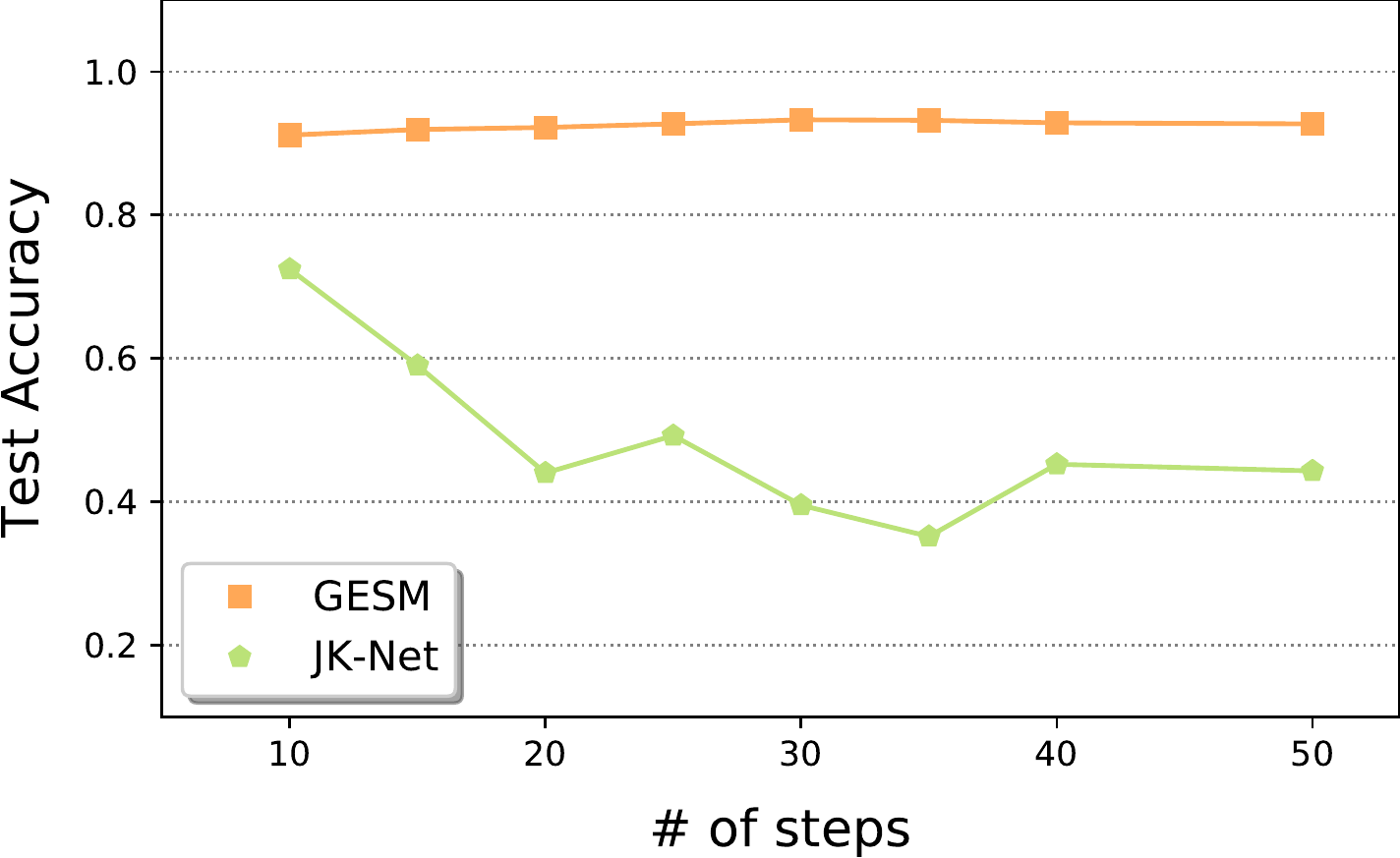}
    \caption{Test accuracy of JK-Net and \prgname{} using concatenated features after the 10th step.}
    \label{fig:test2}
\end{figure}
Interestingly, JK-Net~\citep{xu2018representation} also keeps the training accuracy regardless of the step size by using GCN blocks with multiple steps according to~\figurename~\ref{fig:test1}. We further compared the test accuracy of \prgname{} with JK-Net, a similar approach to our model, in regards to the step size. To investigate the adaptability to larger steps of \prgname{} and JK-Net, we concatenated features after the 10th step. As shown in~\figurename~\ref{fig:test2}, \prgname{} outperforms JK-Net, even though both methods use concatenation to alleviate the oversmoothing issue. These results are in line with the fact that JK-Net obtains global information similar to GCN or GAT. Consequently, the larger the step, the more difficult it is for JK-Net to maintain performance. \prgname{}, on the other hand, maintains a steady performance, which confirms that \prgname{} does not collapse even for large step sizes. 

\begin{figure}
  \centering
  \includegraphics[width=0.75\columnwidth]{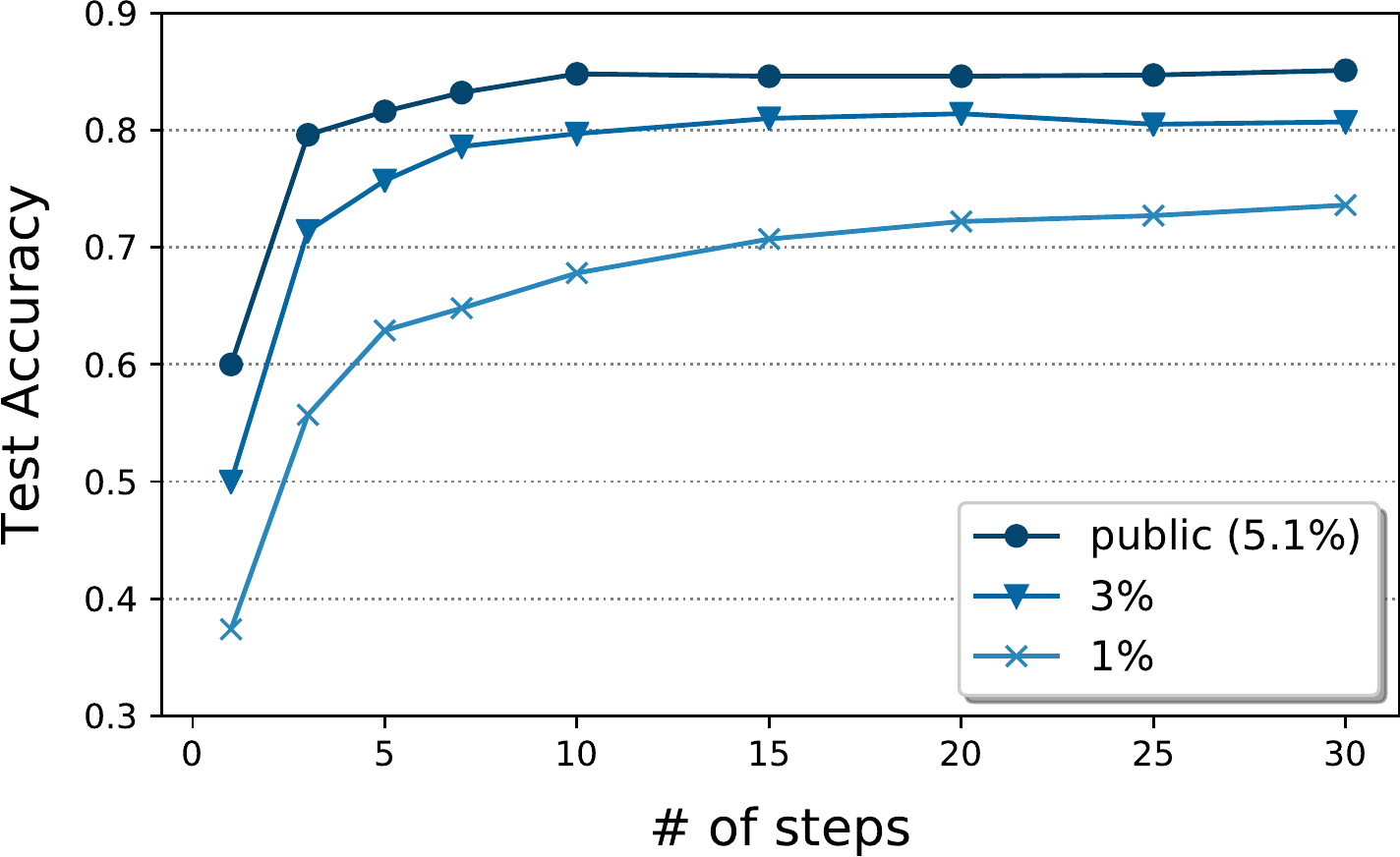}
  \caption{\prgname{} Accuracy comparison for various label rates according to step size.}
  \label{fig:sub22}
\end{figure}
We also observe the effect on accuracy as the number of steps increases under three labeling conditions for \prgname{}. As represented in~\figurename~\ref{fig:sub22}, it is evident that considering remote nodes can contribute to the increase in accuracy. By taking into account more data within a larger neighborhood, our model can make reliable decisions, resulting in improved performance. Inspection of the figure also indicates that the accuracy converges faster for datasets with higher label rates, presumably because a small number of walk steps can be used to explore the entire graph.

\emph{\textbf{Inference time.}}
\begin{figure}
  \centering
  \includegraphics[width=0.75\columnwidth]{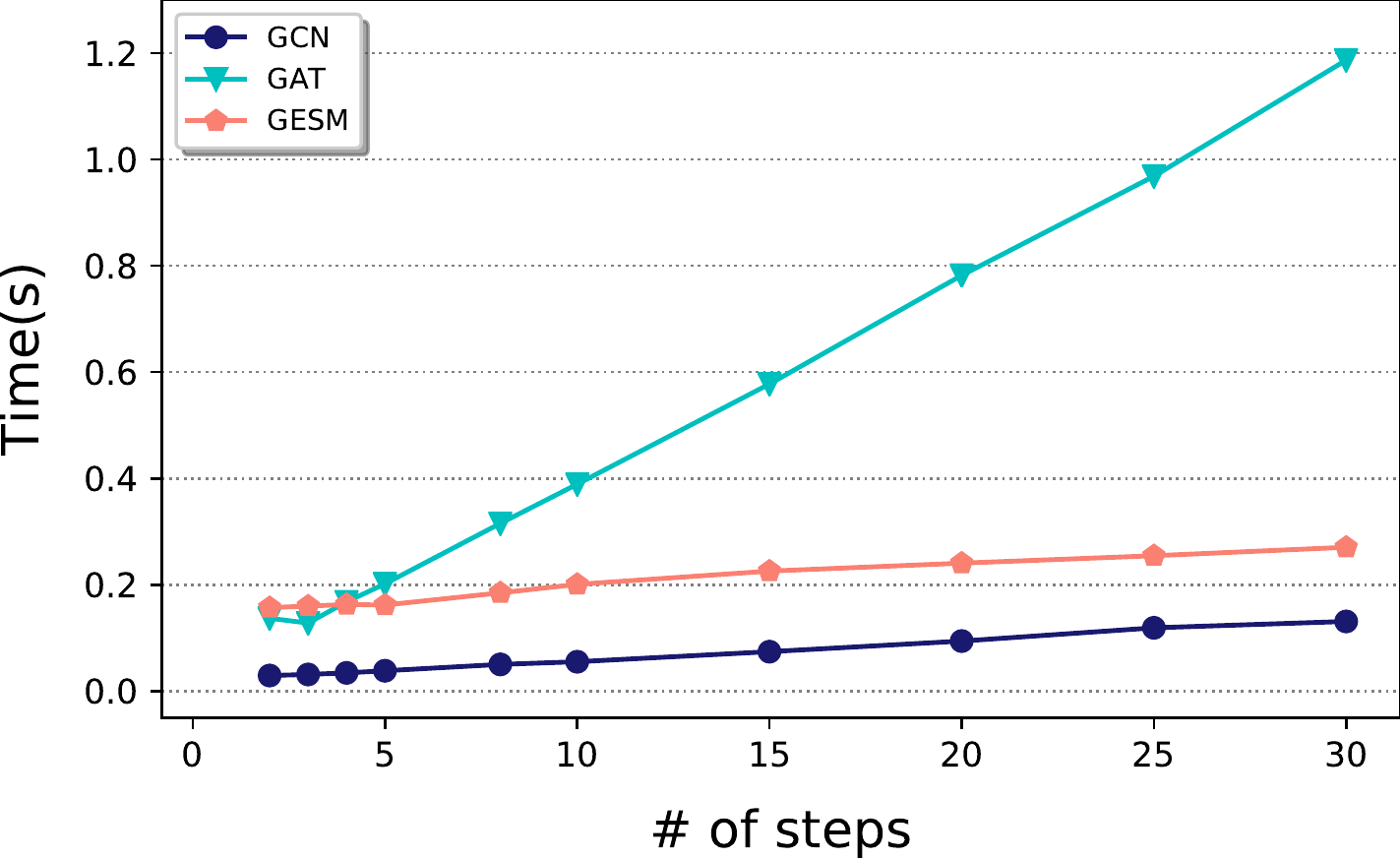}
  \caption{Inference time of various models as the step size increases on Cora.}
  \label{fig:comparison1}
\end{figure}
As shown in \figurename~\ref{fig:comparison1}, the computational complexity of all models increases linearly as the step size increases. We can observe that the inference time of GCN~\citep{kipf2016semi} is slightly faster than that of \prgname{}, especially with a constant margin. The inference time of \prgname{} is much faster than GAT~\citep{velivckovic2017graph} while providing higher accuracies and stable results as shown in~\tablename~\ref{tab:pitfall}. Our methods are both fast and accurate due to the sophisticated design with a mixture of random walk steps.

\emph{\textbf{Embedding visualization.}}
\begin{figure}
    \centering
        \subfigure[\prgname{} (w/o reg)]{\includegraphics[width=40mm,scale=1.0]{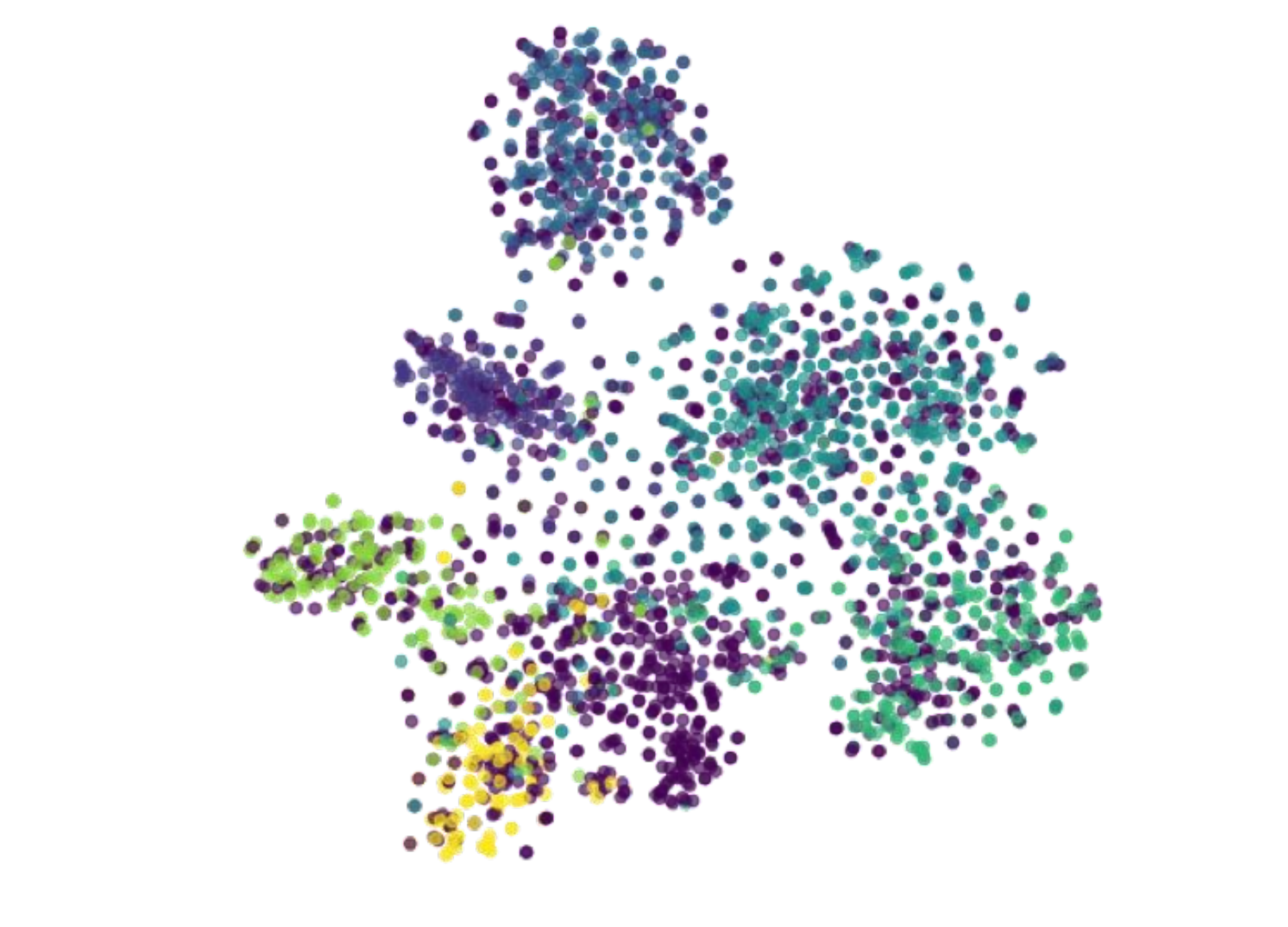}}
        \subfigure[\prgname{}]{\includegraphics[width=40mm,scale=1.0]{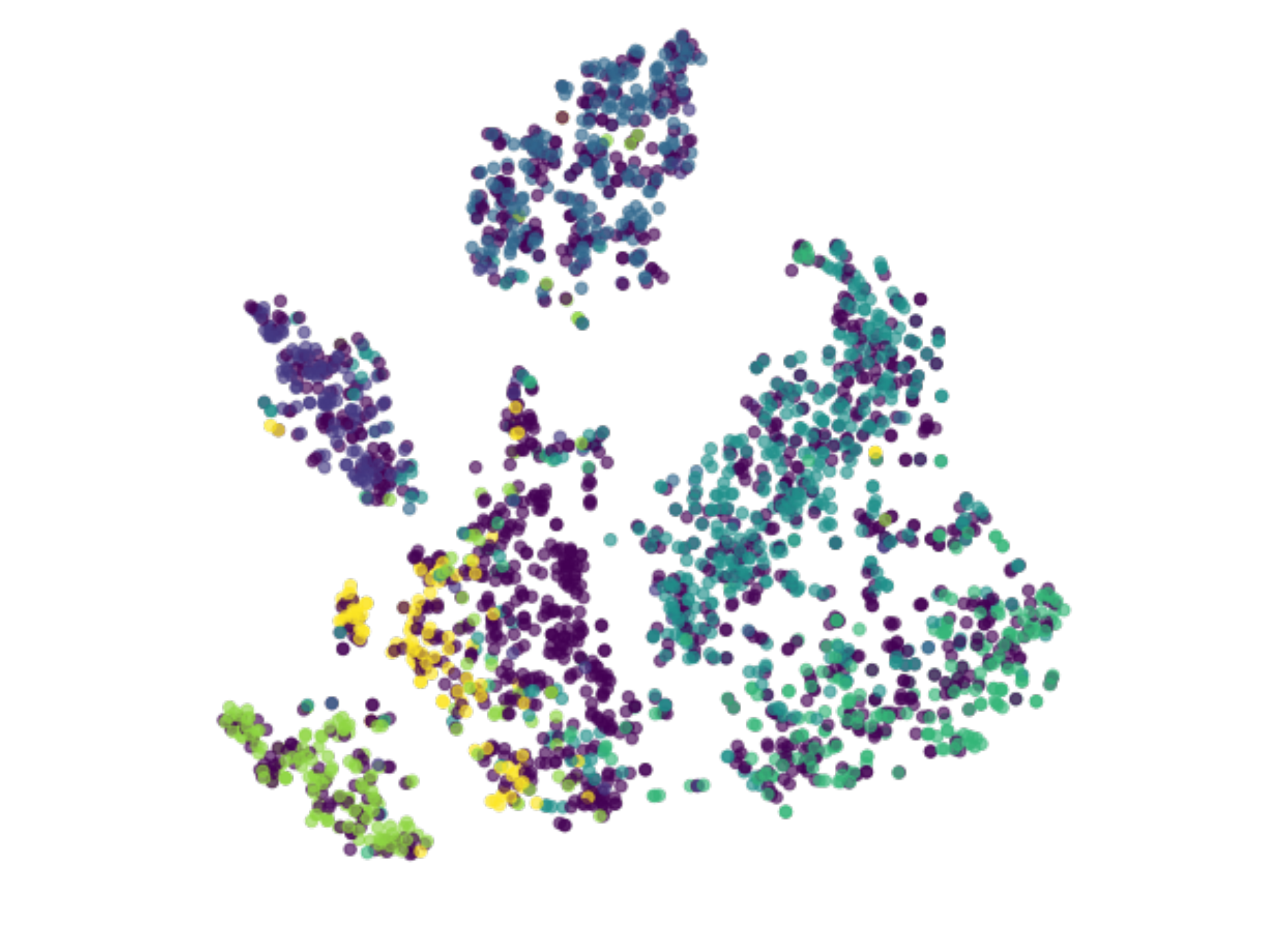}}
    \caption{t-SNE plot of the last hidden layer trained on Cora.}
    \label{fig:embedding}
\end{figure}
\figurename~\ref{fig:embedding} visualizes the hidden features of Cora from our models by using the t-SNE algorithm~\citep{maaten2008visualizing}. The figure illustrates the difference between \prgname{} (w/o reg) and \prgname{}. While the nodes are scattered in the result from \prgname{} (w/o reg), they are closely clustered in that of \prgname{}. According to the results in \tablename~\ref{tab:result}, more closely clustered \prgname{} generally produce better results than loosely clustered \prgname{} (w/o reg).

\section{Conclusion}
Traditional graph neural networks suffer from the oversmoothing issue as increasing propagation steps and poor generalization on unseen graphs. To tackle these issues, we propose a simple but effective model that weights differently depending on node information in the aggregation process and adaptively considers global and local information by employing the mixture of multiple random walk steps. To further refine the graph representation, we have presented a new regularization term, which enforces the similar neighbor nodes to be closely clustered in the node embedding space. The results from extensive experiments show that our \prgname{} successfully achieves state-of-the-art or competitive performance for both transductive and inductive learning tasks including eight benchmark graph datasets.

As future directions, we will refine our method that utilizes node information to improve the computational efficiency regarding attention. In addition, we will extend GESM so that it can be applied to real-world large-scale graph data.

\bibliography{reference}

\begin{thebibliography}{41}
\providecommand{\natexlab}[1]{#1}
\providecommand{\url}[1]{\texttt{#1}}
\expandafter\ifx\csname urlstyle\endcsname\relax
  \providecommand{\doi}[1]{doi: #1}\else
  \providecommand{\doi}{doi: \begingroup \urlstyle{rm}\Url}\fi

\bibitem[Abu-El-Haija et~al.(2019)Abu-El-Haija, Perozzi, Kapoor, Harutyunyan,
  Alipourfard, Lerman, Steeg, and Galstyan]{abu2019mixhop}
Abu-El-Haija, S., Perozzi, B., Kapoor, A., Harutyunyan, H., Alipourfard, N.,
  Lerman, K., Steeg, G.~V., and Galstyan, A.
\newblock Mixhop: Higher-order graph convolution architectures via sparsified
  neighborhood mixing.
\newblock \emph{International Conference on Machine Learning (ICML)}, 2019.

\bibitem[Bahdanau et~al.(2015)Bahdanau, Cho, and Bengio]{bahdanau2014neural}
Bahdanau, D., Cho, K., and Bengio, Y.
\newblock Neural machine translation by jointly learning to align and
  translate.
\newblock \emph{International Conference on Learning Representations (ICLR)},
  2015.

\bibitem[Battaglia et~al.(2018)Battaglia, Hamrick, Bapst, Sanchez-Gonzalez,
  Zambaldi, Malinowski, Tacchetti, Raposo, Santoro, Faulkner,
  et~al.]{battaglia2018relational}
Battaglia, P.~W., Hamrick, J.~B., Bapst, V., Sanchez-Gonzalez, A., Zambaldi,
  V., Malinowski, M., Tacchetti, A., Raposo, D., Santoro, A., Faulkner, R.,
  et~al.
\newblock Relational inductive biases, deep learning, and graph networks.
\newblock \emph{arXiv preprint arXiv:1806.01261}, 2018.

\bibitem[Chami et~al.(2019)Chami, Ying, R{\'e}, and
  Leskovec]{chami2019hyperbolic}
Chami, I., Ying, Z., R{\'e}, C., and Leskovec, J.
\newblock Hyperbolic graph convolutional neural networks.
\newblock In \emph{Advances in Neural Information Processing Systems}, pp.\
  4869--4880, 2019.

\bibitem[Clevert et~al.(2016)Clevert, Unterthiner, and
  Hochreiter]{clevert2015fast}
Clevert, D.-A., Unterthiner, T., and Hochreiter, S.
\newblock Fast and accurate deep network learning by exponential linear units
  (elus).
\newblock \emph{International Conference on Learning Representations (ICLR)},
  2016.

\bibitem[Defferrard et~al.(2016)Defferrard, Bresson, and
  Vandergheynst]{defferrard2016convolutional}
Defferrard, M., Bresson, X., and Vandergheynst, P.
\newblock Convolutional neural networks on graphs with fast localized spectral
  filtering.
\newblock In \emph{Advances in neural information processing systems}, pp.\
  3844--3852, 2016.

\bibitem[Gao \& Ji(2019)Gao and Ji]{gao2019graph}
Gao, H. and Ji, S.
\newblock Graph representation learning via hard and channel-wise attention
  networks.
\newblock In \emph{Proceedings of the 25th ACM SIGKDD International Conference
  on Knowledge Discovery \& Data Mining}, pp.\  741--749. ACM, 2019.

\bibitem[Gao et~al.(2018)Gao, Wang, and Ji]{gao2018large}
Gao, H., Wang, Z., and Ji, S.
\newblock Large-scale learnable graph convolutional networks.
\newblock In \emph{Proceedings of the 24th ACM SIGKDD International Conference
  on Knowledge Discovery \& Data Mining}, pp.\  1416--1424. ACM, 2018.

\bibitem[Gleich \& Kloster(2016)Gleich and Kloster]{gleich2016seeded}
Gleich, D. and Kloster, K.
\newblock Seeded pagerank solution paths.
\newblock \emph{European Journal of Applied Mathematics}, 27\penalty0
  (6):\penalty0 812--845, 2016.

\bibitem[Glorot \& Bengio(2010)Glorot and Bengio]{glorot2010understanding}
Glorot, X. and Bengio, Y.
\newblock Understanding the difficulty of training deep feedforward neural
  networks.
\newblock In \emph{Proceedings of the thirteenth international conference on
  artificial intelligence and statistics}, pp.\  249--256, 2010.

\bibitem[Gordo et~al.(2017)Gordo, Almazan, Revaud, and Larlus]{gordo2017end}
Gordo, A., Almazan, J., Revaud, J., and Larlus, D.
\newblock End-to-end learning of deep visual representations for image
  retrieval.
\newblock \emph{International Journal of Computer Vision}, 124\penalty0
  (2):\penalty0 237--254, 2017.

\bibitem[Hamilton et~al.(2017)Hamilton, Ying, and
  Leskovec]{hamilton2017inductive}
Hamilton, W., Ying, Z., and Leskovec, J.
\newblock Inductive representation learning on large graphs.
\newblock In \emph{Advances in Neural Information Processing Systems}, pp.\
  1024--1034, 2017.

\bibitem[Jia et~al.(2017)Jia, Wang, and Gong]{jia2017random}
Jia, J., Wang, B., and Gong, N.~Z.
\newblock Random walk based fake account detection in online social networks.
\newblock In \emph{2017 47th Annual IEEE/IFIP International Conference on
  Dependable Systems and Networks (DSN)}, pp.\  273--284. IEEE, 2017.

\bibitem[Kim et~al.(2018)Kim, Kim, Seo, Kim, Park, Park, Jo, Kim, Yang, Kim,
  et~al.]{kim2018nsml}
Kim, H., Kim, M., Seo, D., Kim, J., Park, H., Park, S., Jo, H., Kim, K., Yang,
  Y., Kim, Y., et~al.
\newblock Nsml: Meet the mlaas platform with a real-world case study.
\newblock \emph{arXiv preprint arXiv:1810.09957}, 2018.

\bibitem[Kim et~al.(2019)Kim, Kwak, Kwak, Park, Sim, Cho, Kim, Kwon, Sung, and
  Ha]{kim2019tripartite}
Kim, K.-M., Kwak, D., Kwak, H., Park, Y.-J., Sim, S., Cho, J.-H., Kim, M.,
  Kwon, J., Sung, N., and Ha, J.-W.
\newblock Tripartite heterogeneous graph propagation for large-scale social
  recommendation.
\newblock \emph{arXiv preprint arXiv:1908.02569}, 2019.

\bibitem[Kingma \& Ba(2015)Kingma and Ba]{kingma2014adam}
Kingma, D.~P. and Ba, J.
\newblock Adam: A method for stochastic optimization.
\newblock \emph{International Conference on Learning Representations (ICLR)},
  2015.

\bibitem[Kipf \& Welling(2017)Kipf and Welling]{kipf2016semi}
Kipf, T.~N. and Welling, M.
\newblock Semi-supervised classification with graph convolutional networks.
\newblock \emph{International Conference on Learning Representations (ICLR)},
  2017.

\bibitem[Klicpera et~al.(2019)Klicpera, Bojchevski, and
  G{\"u}nnemann]{klicpera2018predict}
Klicpera, J., Bojchevski, A., and G{\"u}nnemann, S.
\newblock Predict then propagate: Graph neural networks meet personalized
  pagerank.
\newblock \emph{International Conference on Learning Representations (ICLR)},
  2019.

\bibitem[Lee et~al.(2018)Lee, Rossi, Kim, Ahmed, and Koh]{lee2018attention}
Lee, J.~B., Rossi, R.~A., Kim, S., Ahmed, N.~K., and Koh, E.
\newblock Attention models in graphs: A survey.
\newblock \emph{arXiv preprint arXiv:1807.07984}, 2018.

\bibitem[Li et~al.(2018)Li, Han, and Wu]{li2018deeper}
Li, Q., Han, Z., and Wu, X.-M.
\newblock Deeper insights into graph convolutional networks for semi-supervised
  learning.
\newblock In \emph{Thirty-Second AAAI Conference on Artificial Intelligence},
  2018.

\bibitem[Liao et~al.(2019)Liao, Zhao, Urtasun, and Zemel]{liao2019lanczosnet}
Liao, R., Zhao, Z., Urtasun, R., and Zemel, R.~S.
\newblock Lanczosnet: Multi-scale deep graph convolutional networks.
\newblock \emph{International Conference on Learning Representations (ICLR)},
  2019.

\bibitem[Luan et~al.(2019)Luan, Zhao, Chang, and Precup]{luan2019break}
Luan, S., Zhao, M., Chang, X.-W., and Precup, D.
\newblock Break the ceiling: Stronger multi-scale deep graph convolutional
  networks.
\newblock \emph{Advances in neural information processing systems}, 2019.

\bibitem[Maaten \& Hinton(2008)Maaten and Hinton]{maaten2008visualizing}
Maaten, L. v.~d. and Hinton, G.
\newblock Visualizing data using t-sne.
\newblock \emph{Journal of machine learning research}, 9\penalty0
  (Nov):\penalty0 2579--2605, 2008.

\bibitem[Manley(2015)]{manley2015estimating}
Manley, E.
\newblock Estimating urban traffic patterns through probabilistic
  interconnectivity of road network junctions.
\newblock \emph{PloS one}, 10\penalty0 (5):\penalty0 e0127095, 2015.

\bibitem[Page et~al.(1999)Page, Brin, Motwani, and Winograd]{page1999pagerank}
Page, L., Brin, S., Motwani, R., and Winograd, T.
\newblock The pagerank citation ranking: Bringing order to the web.
\newblock Technical report, Stanford InfoLab, 1999.

\bibitem[Perozzi et~al.(2014)Perozzi, Al-Rfou, and Skiena]{perozzi2014deepwalk}
Perozzi, B., Al-Rfou, R., and Skiena, S.
\newblock Deepwalk: Online learning of social representations.
\newblock In \emph{Proceedings of the 20th ACM SIGKDD international conference
  on Knowledge discovery and data mining}, pp.\  701--710. ACM, 2014.

\bibitem[Scarselli et~al.(2008)Scarselli, Gori, Tsoi, Hagenbuchner, and
  Monfardini]{scarselli2008graph}
Scarselli, F., Gori, M., Tsoi, A.~C., Hagenbuchner, M., and Monfardini, G.
\newblock The graph neural network model.
\newblock \emph{IEEE Transactions on Neural Networks}, 20\penalty0
  (1):\penalty0 61--80, 2008.

\bibitem[Sen et~al.(2008)Sen, Namata, Bilgic, Getoor, Galligher, and
  Eliassi-Rad]{sen2008collective}
Sen, P., Namata, G., Bilgic, M., Getoor, L., Galligher, B., and Eliassi-Rad, T.
\newblock Collective classification in network data.
\newblock \emph{AI magazine}, 29\penalty0 (3):\penalty0 93--93, 2008.

\bibitem[Shchur et~al.(2018)Shchur, Mumme, Bojchevski, and
  G{\"u}nnemann]{shchur2018pitfalls}
Shchur, O., Mumme, M., Bojchevski, A., and G{\"u}nnemann, S.
\newblock Pitfalls of graph neural network evaluation.
\newblock \emph{arXiv preprint arXiv:1811.05868}, 2018.

\bibitem[Strang(1993)]{strang1993introduction}
Strang, G.
\newblock \emph{Introduction to linear algebra}, volume~3.
\newblock Wellesley-Cambridge Press Wellesley, MA, 1993.

\bibitem[Sung et~al.(2017)Sung, Kim, Jo, Yang, Kim, Lausen, Kim, Lee, Kwak, Ha,
  et~al.]{sung2017nsml}
Sung, N., Kim, M., Jo, H., Yang, Y., Kim, J., Lausen, L., Kim, Y., Lee, G.,
  Kwak, D., Ha, J.-W., et~al.
\newblock Nsml: A machine learning platform that enables you to focus on your
  models.
\newblock \emph{arXiv preprint arXiv:1712.05902}, 2017.

\bibitem[Thekumparampil et~al.(2018)Thekumparampil, Wang, Oh, and
  Li]{thekumparampil2018attention}
Thekumparampil, K.~K., Wang, C., Oh, S., and Li, L.-J.
\newblock Attention-based graph neural network for semi-supervised learning.
\newblock \emph{arXiv preprint arXiv:1803.03735}, 2018.

\bibitem[Veli{\v{c}}kovi{\'c} et~al.(2018)Veli{\v{c}}kovi{\'c}, Cucurull,
  Casanova, Romero, Lio, and Bengio]{velivckovic2017graph}
Veli{\v{c}}kovi{\'c}, P., Cucurull, G., Casanova, A., Romero, A., Lio, P., and
  Bengio, Y.
\newblock Graph attention networks.
\newblock \emph{International Conference on Learning Representations (ICLR)},
  2018.

\bibitem[Wang et~al.(2018)Wang, Wang, Wang, Zhao, Zhang, Zhang, Xie, and
  Guo]{wang2018graphgan}
Wang, H., Wang, J., Wang, J., Zhao, M., Zhang, W., Zhang, F., Xie, X., and Guo,
  M.
\newblock Graphgan: Graph representation learning with generative adversarial
  nets.
\newblock In \emph{Thirty-Second AAAI Conference on Artificial Intelligence},
  2018.

\bibitem[Wu et~al.(2019{\natexlab{a}})Wu, Zhang, Souza~Jr, Fifty, Yu, and
  Weinberger]{wu2019simplifying}
Wu, F., Zhang, T., Souza~Jr, A. H.~d., Fifty, C., Yu, T., and Weinberger, K.~Q.
\newblock Simplifying graph convolutional networks.
\newblock \emph{International Conference Machine Learning (ICML)},
  2019{\natexlab{a}}.

\bibitem[Wu et~al.(2019{\natexlab{b}})Wu, Pan, Chen, Long, Zhang, and
  Yu]{wu2019comprehensive}
Wu, Z., Pan, S., Chen, F., Long, G., Zhang, C., and Yu, P.~S.
\newblock A comprehensive survey on graph neural networks.
\newblock \emph{arXiv preprint arXiv:1901.00596}, 2019{\natexlab{b}}.

\bibitem[Xu et~al.(2019)Xu, Shen, Cao, Qiu, and Cheng]{xu2019graph}
Xu, B., Shen, H., Cao, Q., Qiu, Y., and Cheng, X.
\newblock Graph wavelet neural network.
\newblock \emph{International Conference on Learning Representations (ICLR)},
  2019.

\bibitem[Xu et~al.(2018)Xu, Li, Tian, Sonobe, Kawarabayashi, and
  Jegelka]{xu2018representation}
Xu, K., Li, C., Tian, Y., Sonobe, T., Kawarabayashi, K.-i., and Jegelka, S.
\newblock Representation learning on graphs with jumping knowledge networks.
\newblock \emph{International Conference on Machine Learning (ICML)}, 2018.

\bibitem[Yang et~al.(2016)Yang, Cohen, and Salakhutdinov]{yang2016revisiting}
Yang, Z., Cohen, W.~W., and Salakhutdinov, R.
\newblock Revisiting semi-supervised learning with graph embeddings.
\newblock \emph{International Conference on Machine Learning (ICML)}, 2016.

\bibitem[Yu et~al.(2017)Yu, Yin, and Zhu]{yu2017spatio}
Yu, B., Yin, H., and Zhu, Z.
\newblock Spatio-temporal graph convolutional networks: A deep learning
  framework for traffic forecasting.
\newblock \emph{arXiv preprint arXiv:1709.04875}, 2017.

\bibitem[Zitnik \& Leskovec(2017)Zitnik and Leskovec]{zitnik2017predicting}
Zitnik, M. and Leskovec, J.
\newblock Predicting multicellular function through multi-layer tissue
  networks.
\newblock \emph{Bioinformatics}, 33\penalty0 (14):\penalty0 i190--i198, 2017.

\end{thebibliography}
\bibliographystyle{icml2020}
\end{document}